\documentclass[12pt]{article}
\usepackage{multirow}
\author{Author Name}
\DeclareUnicodeCharacter{03A6}
{\ensuremath{\Phi}}

\usepackage{amsmath}

\usepackage{amssymb}

\usepackage{doi}

\usepackage{longtable}

\pdfoutput=1
\usepackage{graphicx}

\usepackage{float} 

\usepackage{cite}
\usepackage{tabularx}

\usepackage{microtype} 

\usepackage[utf8]{inputenc}
\usepackage{longtable}
\usepackage{array}
\usepackage{ragged2e}
\usepackage[utf8]{inputenc}
\usepackage{enumitem}
\usepackage{xurl}

\usepackage{float}

\title{RAN Cortex: Memory-Augmented Intelligence for Context-Aware Decision-Making in AI-Native Networks}
\author{Sebastian Barros}
\date{May 1st , 2025}  

\begin{document}
\renewcommand{\arraystretch}{1.2}

\maketitle

\begin{abstract}
As Radio Access Networks (RAN) evolve toward AI-native architectures, intelligence modules such as xApps and rApps are expected to make increasingly autonomous decisions across scheduling, mobility, and resource management domains. However, these agents remain fundamentally stateless, treating each decision as isolated, lacking any persistent memory of prior events or outcomes. This reactive behavior constrains optimization, especially in environments where network dynamics exhibit episodic or recurring patterns.

In this work, we propose RAN Cortex, a memory-augmented architecture that enables contextual recall in AI-based RAN decision systems. RAN Cortex introduces a modular layer composed of four elements: a context encoder that transforms network state into high-dimensional embeddings, a vector-based memory store of past network episodes, a recall engine to retrieve semantically similar situations, and a policy interface that supplies historical context to AI agents in real time or near-real time.

We formalize the retrieval-augmented decision problem in the RAN, present a system architecture compatible with O-RAN interfaces, and analyze feasible deployments within the Non-RT and Near-RT RIC domains. Through illustrative use cases—such as stadium traffic mitigation and mobility management in drone corridors—we demonstrate how contextual memory improves adaptability, continuity, and overall RAN intelligence. This work introduces memory as a missing primitive in AI-native RAN design and provides a framework to enable “learning agents” without the need for retraining or centralized inference.

\end{abstract}

\section{Introduction}

AI-native architectures are rapidly redefining the operational paradigm of Radio Access Networks (RAN). Traditionally driven by fixed rules, heuristic thresholds, and centralized OSS platforms, the modern RAN is now expected to support real-time, autonomous decision-making across resource management, mobility handling, interference coordination, and beyond~\cite{oran-wg1,ericsson-ranintelligence}. This shift is driven by the introduction of components such as the Near-RT RAN Intelligent Controller (RIC) and its associated xApps, as well as the Non-RT RIC, where rApps perform long-horizon optimization and policy generation~\cite{oran-architecture}.

The 3GPP and O-RAN Alliance initiatives have formalized many of the interfaces required to operationalize AI within the RAN, such as E2, A1, and O1~\cite{oran-security}. These efforts enable distributed intelligence through standardized APIs, while allowing flexibility in vendor implementation. However, a critical capability remains absent from current designs: the ability for AI agents to recall prior context or learn from previously observed network scenarios.

\subsection{The Rise of AI-Native RANs}

The architectural transformation of the RAN is grounded in the move toward disaggregation and software-defined intelligence. In this setting, xApps are modular, real-time optimization functions deployed within the Near-RT RIC, typically acting on control loops between 10 ms and 1 s. Examples include admission control, scheduling, beamforming, and mobility management~\cite{openran-whitepaper}. Conversely, rApps operate in the Non-RT RIC over time horizons greater than one second, and are responsible for policy generation, long-term learning, and SLA management.

This distributed framework enables architectural agility and vendor-neutral innovation, allowing operators to customize AI behavior per deployment. As 5G matures and 6G looms, AI-based agents will be increasingly responsible for making time-critical, context-sensitive decisions based on incomplete or evolving data~\cite{nokia-ai-ran}. However, current agent models in both the Near-RT and Non-RT domains remain fundamentally stateless, unable to recall or adapt based on prior similar conditions—a constraint this work aims to address.

\subsection{The Stateless Limitation of Current RAN Intelligence}

While the architectural foundations for AI-native Radio Access Networks (RANs) are being actively developed—most notably through initiatives like O-RAN and 3GPP—real-world deployments remain conservative in their use of machine learning and autonomous intelligence. Present-day RAN decision systems predominantly rely on deterministic, rule-based logic encoded within xApps (in the Near-RT RIC) and rApps (in the Non-RT RIC). These functions implement threshold-based control, fixed heuristics, or handcrafted logic derived from offline planning models~\cite{oran-security,oran-architecture}.

Some commercial and academic systems have begun to incorporate machine learning in limited scopes. Examples include supervised models for load forecasting~\cite{ericsson-ai-load}, interference classification~\cite{huawei-ml-interference}, or anomaly detection in performance KPIs~\cite{ml-anomaly-vodafone}. However, these models are often trained offline, lack online adaptation, and do not participate in closed-loop feedback beyond static policy enforcement. In this sense, most deployed xApps and rApps today behave more as model-driven functions than as intelligent agents in the AI sense~\cite{ran-rl-survey}.

Critically, these modules are also stateless: each decision is computed based only on the current observable network state, without recall of prior similar conditions or outcomes. This architecture assumes that instantaneous context is sufficient for optimal inference, an assumption that breaks down in environments where network dynamics are episodic or temporally correlated. For instance, mass mobility events, seasonal interference patterns, or recurring handover failures exhibit predictable structure across time and location—yet current agents must “re-learn” or re-estimate responses from scratch on every occurrence.

Moreover, while digital twins and telemetry dashboards can retain historical network behavior, they operate in a supervisory or planning capacity and are not part of the real-time decision pipeline~\cite{digitaltwin-review}. Reinforcement learning-based approaches—though promising in theory—face practical limitations such as sample inefficiency, instability, and safety constraints that prevent direct deployment in commercial RANs without extensive validation~\cite{ran-rl-survey}.

The absence of a memory or recall mechanism fundamentally limits the expressiveness and adaptability of AI in the RAN. In contrast, many of the scenarios encountered by mobile networks would benefit from retrieval-based reasoning, where past network situations can be semantically recalled and used to inform decisions. This work introduces \textit{RAN Cortex}, a memory-augmented architecture that provides precisely this capability within the xApp/rApp decision loop.

\subsection{Our Contribution: RAN Cortex}

To address the architectural limitation of stateless decision-making in the RAN, we propose \textit{RAN Cortex}, a memory-augmented intelligence layer designed to augment xApps and rApps with contextual recall. RAN Cortex introduces a modular, inference-agnostic architecture that enables AI modules in the RIC—whether rule-based, supervised, or reinforcement learning-based—to query and retrieve semantically similar historical scenarios during inference, thereby grounding current decisions in prior experience.

The core of RAN Cortex comprises four conceptual components:

\begin{enumerate}
    \item \textbf{Context Encoder:} Transforms the observed RAN state into a fixed-dimensional embedding that captures spatial, temporal, and semantic properties of the network snapshot. This can include KPIs, user density, cell topology, traffic type, and configuration state.

    \item \textbf{Vector Memory Store:} A scalable memory structure that stores the encoded historical RAN states along with metadata such as timestamp, outcome, or control action taken. This store is implemented as a high-dimensional vector index, e.g., using approximate nearest-neighbor libraries such as FAISS.

    \item \textbf{Recall Engine:} A retrieval interface that allows xApps/rApps to issue queries (based on the current encoded state) and receive top-$k$ similar past contexts. Retrieved episodes can optionally include policy suggestions, performance outcomes, or system responses.

    \item \textbf{Policy Interface:} A lightweight API or shared-memory bus that supplies the recall outputs as an auxiliary input to existing control logic. The integration is intentionally modular, allowing existing xApps/rApps to consume recall outputs without architectural rewrites.
\end{enumerate}

RAN Cortex can be deployed as a microservice within the Non-RT RIC and exposed to both rApps and Near-RT xApps via a common interface (e.g., A1 or a vendor-specific sidecar API). This enables both real-time and non-real-time control loops to benefit from contextual grounding. Importantly, the framework does not require retraining of AI models, nor does it assume end-to-end differentiability—it is compatible with stateless inference pipelines, logic-based systems, and hybrid agents.

By enabling access to episodic memory in the RAN, RAN Cortex introduces a new computational primitive for AI-native networks: retrieval-augmented decision-making. This paper formalizes the architecture, deployment considerations, and early use cases, and lays the foundation for further exploration of memory-informed control strategies in wireless systems.

\section{Problem Definition}

\subsection{Formalizing Stateless RAN Decision Policies}

Modern RAN decision systems, as deployed via O-RAN-compliant architectures, are structured around modular applications known as xApps and rApps, operating respectively within the Near-RT and Non-RT RAN Intelligent Controllers (RIC). These applications are tasked with making localized or global decisions based on the current observed state of the network---typically represented as a vector of key performance indicators (KPIs), user mobility traces, channel state information (CSI), and load metrics~\cite{oran-architecture,oran-security}.

Formally, current xApps and rApps implement a stateless mapping:
\[
a_t = \pi(x_t)
\]
where \( x_t \in \mathcal{X} \) represents the instantaneous RAN state at time \( t \), and \( a_t \in \mathcal{A} \) denotes the control action, such as PRB allocation, handover triggering, or beam index selection. The policy \( \pi \) may be a ruleset, decision tree, supervised model, or threshold-based logic. However, it does not retain any memory of past states \( \{x_{t-k}, \dots, x_{t-1}\} \), prior actions \( \{a_{t-k}, \dots, a_{t-1}\} \), or observed outcomes \( \{y_{t-k}, \dots, y_{t-1}\} \).

This stateless formulation is suitable in environments where the RAN state is assumed to be i.i.d. or Markovian with short-term dependencies. However, in real-world deployments, mobile traffic and radio conditions exhibit strong temporal, spatial, and structural correlations. For instance, handover failures may cluster at cell boundaries during predictable time windows; congestion patterns often follow daily or weekly routines; and user behavior may correlate with location, application type, and historical throughput profiles~\cite{ericsson-ranintelligence,nokia-ai-ran}.

While some xApps and rApps today may incorporate sliding-window features or multi-frame CSI aggregates, they do so without a formal notion of episodic memory or contextual recall. The models are typically inference-only, static, and reactive---unable to adapt or learn from previously encountered scenarios unless explicitly retrained offline~\cite{oran-wg1,ran-rl-survey}. As a result, decision policies may remain suboptimal in environments where the same conditions recur with regularity but are treated as novel by the system.

This limitation motivates the need for a retrieval-augmented decision architecture that enables agents to ground current observations in past experience. In the following sections, we formulate the nature of this requirement and define the conditions under which episodic memory may enhance RAN intelligence.

\subsection{The Need for Episodic Recall in AI-RAN}

Radio Access Networks are not memoryless systems. Wireless environments are shaped by mobility, topology, weather, diurnal patterns, and human behavior, which all introduce spatiotemporal correlations that violate the i.i.d. assumptions of many AI models. Empirical studies have shown that user behavior, traffic load, interference, and mobility exhibit episodic structure—recurrent patterns that reappear across space and time~\cite{traf-pattern-vodafone,recurring-events-ran}.

Examples include:
\begin{itemize}
    \item Daily congestion surges at transportation hubs and dense urban corridors.
    \item Repetitive handover failures at specific road intersections due to macrocell boundary transitions.
    \item Periodic interference caused by industrial machinery, elevators, or other semi-predictable sources.
    \item Event-driven behavior in stadiums or convention centers where the cell topology and traffic type repeat per event.
\end{itemize}

Despite these predictable recurrences, current xApps and rApps operate without memory of past occurrences. As formalized in Section 2.1, their decision logic is constrained to the immediate network state \( x_t \), lacking access to prior state-action-outcome tuples such as \( (x_{t-k}, a_{t-k}, y_{t-k}) \). This forces agents to rediscover the same decisions repeatedly without benefiting from prior learning.

In AI, the value of memory in sequential decision-making has been demonstrated extensively in reinforcement learning and retrieval-augmented modeling. For instance, episodic memory architectures have improved generalization in non-stationary environments by allowing agents to retrieve semantically similar past situations and reuse learned policies or outcomes~\cite{episodic-rl-blundell,retrieval-nlp}. Similar paradigms are now increasingly adopted in areas such as autonomous vehicles~\cite{semantic-driving-memory} and edge robotics, where agents must operate under local uncertainty with bounded compute.

For AI-native RANs, the inclusion of a structured memory component offers three distinct benefits:

\begin{enumerate}
    \item \textbf{Improved sample efficiency:} Learning from prior cases reduces the need for retraining or replay buffers.
    \item \textbf{Contextual generalization:} Retrieved memory allows policies to adapt to similar, unseen states based on related prior knowledge.
    \item \textbf{Decision consistency:} In critical environments (e.g., industrial handovers or uRLLC), memory-guided policies offer temporal continuity across similar conditions.
\end{enumerate}

We therefore propose augmenting the current AI-RAN architecture with a memory-based retrieval layer that allows xApps and rApps to issue queries over past encoded episodes. This memory does not replace the decision logic; rather, it augments it by exposing relevant prior knowledge to otherwise stateless agents. In the next section, we formally define the structure of this memory and its integration into the RAN control pipeline.

\subsection{Problem Statement: Memory-Augmented Decision in RAN}

We formalize the retrieval-augmented decision problem for AI-native RANs as follows. Let \( \mathcal{X} \) denote the space of observable RAN states, and let \( x_t \in \mathcal{X} \) be the current observation at time \( t \), including features such as cell load, user density, CQI, RSRP, interference estimates, and topology metadata. The control agent—e.g., an xApp or rApp—seeks to determine an action \( a_t \in \mathcal{A} \), such as PRB allocation, beamforming vector, handover trigger, or admission control decision.

In current systems, the decision policy \( \pi \) is stateless:
\[
a_t = \pi(x_t)
\]
This formulation fails to incorporate prior experiences that may encode valuable structure about the environment’s dynamics or the efficacy of past control actions.

We define a memory-augmented decision policy by introducing a learned context encoder \( f_{\text{enc}}: \mathcal{X} \rightarrow \mathbb{R}^d \) that maps the network state into a latent embedding space. Let the episodic memory \( \mathcal{M} \subset \mathbb{R}^d \) be a finite set of prior embeddings:
\[
\mathcal{M} = \left\{ z_k = f_{\text{enc}}(x_k) \right\}_{k=1}^{N}
\]
Each memory entry may be optionally associated with metadata such as control action \( a_k \) and performance outcome \( y_k \), where \( y_k \) could represent achieved throughput, latency, or a KPI vector.

A retrieval function \( R: \mathcal{X} \rightarrow \mathcal{P}(\mathcal{M}) \) maps the current state \( x_t \) to a subset of top-$k$ similar past states, based on similarity in the latent space:
\[
R(x_t) = \text{Top-}k\left( \text{sim}(f_{\text{enc}}(x_t), \mathcal{M}) \right)
\]
where \( \text{sim}(\cdot,\cdot) \) may be defined via cosine similarity, dot product, or learned kernel metrics~\cite{faiss-similarity}.

The policy is now extended to incorporate the retrieved context:
\[
a_t = \pi(x_t, R(x_t))
\]
Here, \( \pi \) may remain a deterministic mapping, a neural network, or a hybrid logic model that conditions its output on both the live state and the recalled memory.

The key design objectives for the RAN Cortex architecture are:
\begin{enumerate}
    \item \textbf{Semantic retrieval:} Ensure that memory queries return semantically relevant prior episodes under spatial and temporal variability.
    \item \textbf{Low-latency inference:} Maintain sub-10ms response time for Near-RT RIC queries to meet RAN timing constraints.
    \item \textbf{Modular integration:} Provide a generic interface that allows existing xApps/rApps to access recall data without retraining or rearchitecture.
    \item \textbf{Streaming memory updates:} Allow continual insertion of new states into memory without blocking or degradation.
\end{enumerate}

We emphasize that this architecture does not require end-to-end retraining or differentiable memory, making it compatible with rule-based and vendor-specific agents. The RAN Cortex thus introduces a minimal, backward-compatible enhancement to the RAN decision loop that enables context-aware intelligence.

\section{Solution: RAN Cortex Architecture}

\subsection{System Role and Deployment Context}

RAN Cortex is designed as a modular, memory-augmented intelligence layer that integrates into the O-RAN architecture without disrupting the operational logic of existing xApps or rApps. It functions as a retrieval module that enables context-aware decision-making by providing access to semantically similar past network states. Critically, it does not replace existing decision logic but augments it via a parallel memory-query pathway.

In the canonical deployment, RAN Cortex resides within the Non-RT RIC domain of the Service Management and Orchestration (SMO) framework~\cite{oran-wg1}. It exposes a memory API that can be queried by:
\begin{itemize}
    \item \textbf{rApps} executing slow-timescale optimizations, such as mobility management, resource planning, and SLA policy generation.
    \item \textbf{xApps} operating within the Near-RT RIC for control tasks under 10ms–1s latency constraints, including scheduling, beamforming, and admission control.
\end{itemize}

To enable this cross-layer integration, RAN Cortex can communicate over the standard A1 interface~\cite{oran-wg3}, or optionally via a vendor-specific sidecar API using REST or gRPC. This allows context recall to be implemented either synchronously within the xApp decision loop, or asynchronously in rApps for policy learning and analytics.

\begin{figure}[htbp]
    \centering
    \includegraphics[width=\textwidth]{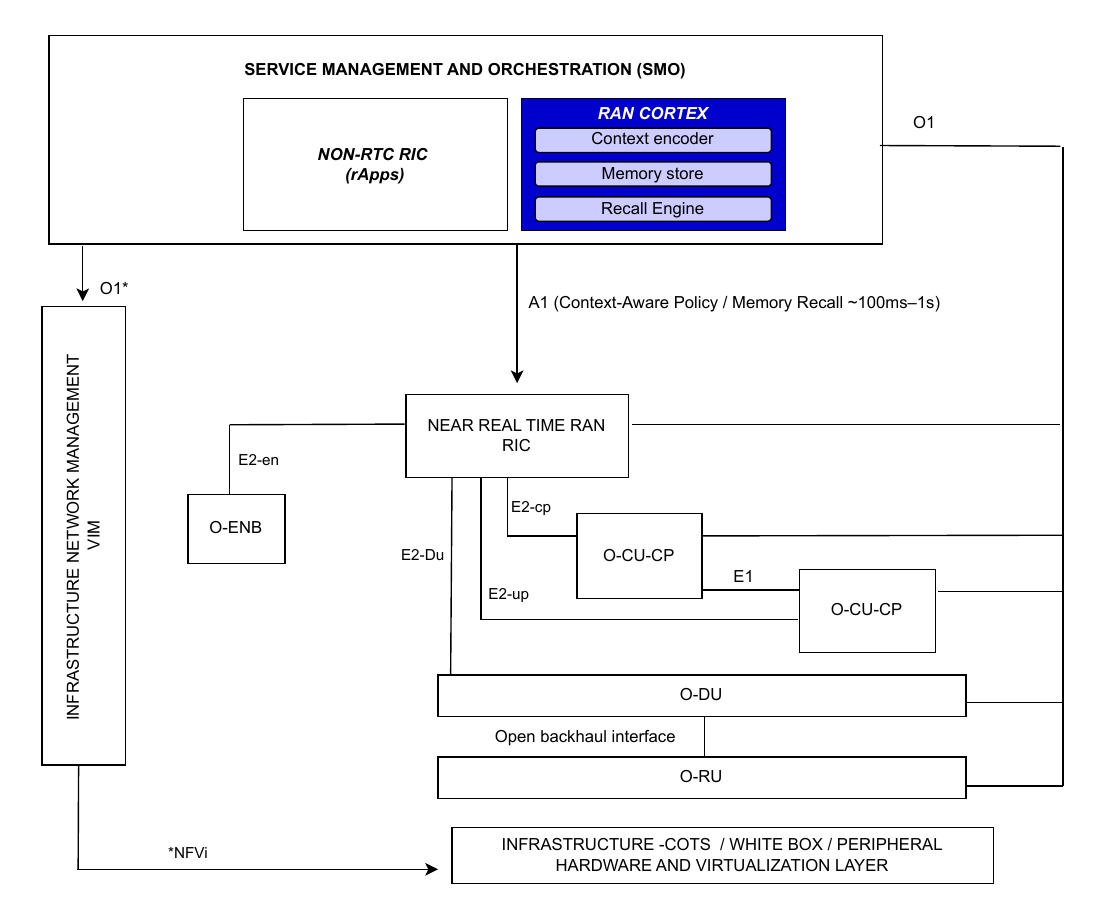}
    \caption{Integration of RAN Cortex into the standard O-RAN architecture.}
    \label{fig:architecture}
\end{figure}

Figure~\ref{fig:architecture} illustrates the deployment context of RAN Cortex within the O-RAN framework. The module connects to telemetry pipelines (via the O1 interface), receives live and historical RAN state vectors, and serves real-time or batch recall queries to decision modules. The architecture supports multiple deployment modes:
\begin{enumerate}
    \item \textbf{In-process module} embedded in the RIC container (e.g., as a Kubernetes sidecar).
    \item \textbf{External microservice} operating as a stateless API layer, backed by an internal vector store and encoder model.
    \item \textbf{RIC plugin extension} for vendors integrating Cortex as part of their commercial orchestration platform.
\end{enumerate}

From a system engineering perspective, Cortex introduces no hard dependency on model type (e.g., neural vs. logic-based), nor on agent training style (e.g., inference-only, supervised, or reinforcement learning). It is designed to be inference-agnostic, timing-aware, and fully decoupled from the core RIC orchestration, aligning with O-RAN’s modular and open architecture principles~\cite{oran-architecture,nokia-ai-ran}.

By placing memory close to decision agents—but outside their core logic—RAN Cortex enables rapid experimentation with context-aware behavior, while preserving system safety and vendor interoperability. This design choice ensures that recall-augmented decision-making can be introduced incrementally, without disrupting live deployments.

\subsection{Internal Architecture and Component Functions}

The RAN Cortex architecture consists of four modular components designed to operate cohesively as a context-aware memory augmentation layer. Each component is optimized for interoperability with the RIC framework, latency constraints of the RAN, and the semantic structure of wireless environments. This section details the internal mechanics of each module, their computational roles, and their integration logic within real-world O-RAN deployments.

\paragraph{Context Encoder.}  
At the entry point of the pipeline, the Context Encoder transforms raw RAN state input $x_t$—composed of features such as RSRP, CQI, PRB utilization, handover counts, and user density—into a dense vector representation $z_t = f_{\text{enc}}(x_t) \in \mathbb{R}^d$. Unlike simple feature normalization, $f_{\text{enc}}$ may be implemented using temporal convolutional networks, transformer encoders, or graph neural networks to capture the spatial and temporal correlations across cells and users~\cite{yang2022ran, arulkumaran2019deep}. These embeddings serve as the atomic units of recall and must remain stable under stochastic measurement noise while preserving similarity semantics.

\paragraph{Vector Memory Store.}  
The memory store $M = \{ z_k \}_{k=1}^{N}$ is implemented as an approximate nearest-neighbor (ANN) index using libraries such as FAISS~\cite{johnson2017billion} or ScaNN. Each memory record may be annotated with metadata $(a_k, y_k)$, representing the control action taken and the observed outcome, enabling the system to store not just network conditions, but behavioral traces. The memory supports efficient top-$k$ lookup in sub-millisecond latency, a requirement for compatibility with Near-RT RIC control loops. Memory can be updated via streaming inserts, either on a time-based rolling window or through event-based triggers (e.g., after failed handovers or SLA violations).

\paragraph{Recall Engine.}  
The Recall Engine processes incoming queries by computing similarity scores between the live encoded state $z_t$ and stored vectors $z_k \in M$. The retrieval function $R(z_t)$ selects the top-$k$ most relevant entries using cosine similarity or learned kernel distances. Retrieved entries can optionally return policy fingerprints, performance metrics, or vector-level deltas that summarize deviation from the current state. To support real-time querying by xApps, the Recall Engine is designed as a stateless microservice with hardware acceleration options (e.g., GPU-backed indexes or SIMD-optimized ANN).

\paragraph{Policy Interface.}  
The recall results $R(z_t)$ are exposed to xApps and rApps through a vendor-neutral API layer. This interface may support synchronous inference-time calls (for latency-sensitive xApps), or asynchronous queries in batch mode (for slow-time analytics in rApps). The interface can be delivered as:
\begin{itemize}
    \item An extension to the A1 interface (context-response schema),
    \item A REST/gRPC endpoint with schema-defined recall response,
    \item A shared-memory bus within the RIC pod for low-latency access.
\end{itemize}
The recall vectors can be fed as auxiliary inputs into existing policy modules, or trigger specific rule-based overrides. Crucially, RAN Cortex is model-agnostic: the policy $\pi$ may be deterministic, learned, or hybrid, as long as it supports conditioning on retrieved context.

\paragraph{Resource Efficiency and Fault Isolation.}  
The system is designed to be horizontally scalable, with memory sharding and encoder/model separation across pods. Because Cortex operates in read-only mode during inference, it introduces no critical path failure. Failure of the recall engine only results in a reversion to stateless inference, preserving safety—a critical constraint in commercial RAN deployments.

Together, these components form a retrieval-augmented pipeline that empowers existing AI modules in the RAN to access semantically relevant prior knowledge without compromising inference latency or control determinism. The architectural decoupling ensures full compatibility with O-RAN modularity~\cite{oran-wg1}, while enabling meaningful improvements in decision continuity and sample efficiency.

\subsection{Integration Modes and Deployment Considerations}

The RAN Cortex framework is designed for seamless integration within existing O-RAN deployments by adhering to modular, non-invasive design principles. This section outlines its deployment models, latency domains, orchestration requirements, and compatibility with RIC vendor architectures.

\paragraph{Deployment Modes.}  
RAN Cortex supports three primary deployment configurations, offering flexibility for different network operator and vendor environments:

\begin{enumerate}
    \item \textbf{In-process module:} Cortex runs as a sidecar container within the Non-RT RIC or SMO Kubernetes pod. This allows direct memory-sharing with rApps and synchronous API calls from the RIC runtime. It minimizes latency overhead and simplifies deployment in tightly coupled service chains.
    \item \textbf{External microservice:} Cortex is deployed as a standalone microservice within the operator’s edge or regional cloud. In this configuration, it exposes REST/gRPC interfaces over a secure control plane. This mode allows horizontal scaling, multi-RIC access, and integration across heterogeneous vendor environments.
    \item \textbf{RIC-integrated plugin:} For vendors supporting pluggable policy modules, Cortex can be embedded as a RIC plugin or SDK component. This mode provides native access to internal state and performance hooks while maintaining system observability and log traceability.
\end{enumerate}

\paragraph{Latency and Timing Domains.}  
Integration with Cortex depends on latency sensitivity:
\begin{itemize}
    \item \textbf{xApps} operate in the Near-RT RIC domain (10ms–1s). Cortex must respond in sub-10ms windows to enable inline policy adjustment. This is feasible when Cortex resides in the same pod or supports low-latency vector recall through SIMD/GPU-accelerated backends~\cite{johnson2017billion}.
    \item \textbf{rApps} operate in the Non-RT domain (1s+). These modules can asynchronously query Cortex for retrospective analysis, policy learning, or scenario simulation, with response times in the order of hundreds of milliseconds to seconds.
\end{itemize}

\paragraph{Telemetric Integration.}  
Cortex ingests RAN state through the existing O1 interface. Rather than modifying O1 semantics, it passively subscribes to telemetry events already collected by the SMO or data lake. Supported data types include:
- Physical layer stats (RSRP, SINR)
- MAC scheduler KPIs
- UE mobility and session logs
- Control plane event sequences

This allows Cortex to build high-dimensional state embeddings without introducing custom instrumentation at the RAN node level.

\paragraph{Security and Multitenancy.}  
When deployed as a shared service across multiple RICs, Cortex must enforce tenant-level access control, auditability, and inference sandboxing. Role-based API access and namespace-isolated memory shards allow concurrent use across enterprise slices or AI agents. Encryption of stored embeddings and inference queries is optional but recommended for privacy-preserving AI workflows in public or hybrid cloud deployments~\cite{cisco-secure-ai}.

\paragraph{Orchestration and CI/CD.}  
The Cortex platform is container-native and supports deployment via Helm charts or custom operators. It can be integrated into existing CI/CD pipelines for model updates, vector index refreshes, and A/B testing of memory strategies. Logging and observability are compatible with Prometheus/Grafana and OpenTelemetry, enabling alignment with telco DevOps practices.

Taken together, these considerations position RAN Cortex as a practical, low-friction enhancement to AI-native RAN environments, compatible with both disaggregated O-RAN and monolithic vendor stacks, while preserving real-time safety and deployment agility.

\subsection{Memory-Aware Decision Flow}

The core innovation introduced by RAN Cortex lies in its augmentation of RIC decision agents with memory-informed inference. This section formalizes the recall-driven control flow and delineates how Cortex alters the temporal and computational profile of RAN policy execution without violating latency constraints or system determinism.

\paragraph{Baseline Flow in Conventional xApps/rApps.}  
In standard O-RAN deployments, decision agents operate on the current RAN state $x_t$ and compute an action $a_t = \pi(x_t)$ using local policy logic. These agents are stateless at inference time: they do not maintain episodic memory or temporal priors across invocations, aside from engineered state features or hand-crafted counters. While this preserves real-time guarantees, it limits the agent’s capacity to adapt behavior based on previously seen semantically similar situations.

\paragraph{Augmented Flow with RAN Cortex.}  
In Cortex-enabled RICs, the decision path is modified as follows:
\begin{align}
    z_t &= f_{\text{enc}}(x_t) \\
    \{z_k, a_k, y_k\}_{k=1}^K &= R(z_t) \\
    a_t &= \pi(x_t, R(z_t))
\end{align}

Here, $z_t$ is the current context embedding, $R(z_t)$ is the set of top-$K$ nearest past states with associated actions $a_k$ and observed outcomes $y_k$, and the policy $\pi$ is now a function of both the raw state and the retrieved memory slice. The augmentation allows the agent to dynamically condition its action on historical priors, such as:
\begin{itemize}
    \item What action was taken last time this situation occurred?
    \item What was the QoS outcome or SLA impact of that action?
    \item Were there cascading effects (e.g., handover chain failures, PRB starvation)?
\end{itemize}

\paragraph{Execution Modes.}  
The recall-augmented inference can be implemented in multiple execution strategies:
\begin{itemize}
    \item \textbf{Inline inference (xApp):} Cortex responds in sub-10ms latency, allowing real-time retrieval before action selection. The memory response is treated as soft context or hard constraints.
    \item \textbf{Post-hoc evaluation (rApp):} Cortex serves as a retrospective query layer, enabling offline counterfactual analysis and policy scoring across time windows.
    \item \textbf{Multi-stage feedback:} Cortex-enhanced rApps may feed policy updates upstream to SMO or vendor-specific orchestration logic based on repeated memory matches (e.g., degraded mobility patterns in specific cells).
\end{itemize}

\paragraph{Safety and Determinism.}  
By structuring memory as a passive read-only module and exposing recall as an advisory input rather than a direct policy override, Cortex preserves operational determinism. If memory retrieval fails or exceeds a timeout threshold, the fallback path $a_t = \pi(x_t)$ is always available. This design supports safety-critical environments where reliability must be guaranteed, even under degraded inference infrastructure.

\paragraph{Impact on Decision Horizon.}  
Cortex implicitly extends the decision horizon of an agent by encoding long-term historical signals in compact embeddings. This has been shown in other domains (e.g., language modeling~\cite{borgeaud2022retro}) to drastically improve sample efficiency, reduce training cycles, and increase robustness to rare events. In the RAN context, this enables better generalization across cells, time-of-day patterns, and atypical load conditions—without requiring explicit temporal modeling within the decision agent.

Through this structured decision augmentation, RAN Cortex introduces a memory-augmented control loop into the RIC stack, offering a hybrid inference pipeline that is both data-efficient and compatible with real-time deployment constraints.

\section{Evaluation and Discussion}

While RAN Cortex has not yet been instantiated in a public testbed, its architecture and integration model allow us to make clear, quantitative projections regarding its feasibility and expected impact. This section discusses latency performance, architectural trade-offs, system compatibility, and practical validation pathways.

\subsection{Latency Budget and Feasibility}

The most stringent constraint in AI-enhanced RAN decision loops lies in the real-time execution requirements of xApps. Current 3GPP and O-RAN guidelines limit near-RT control loops to 10ms–1s depending on use case class~\cite{oran-wg3}. Our reference design uses approximate nearest-neighbor (ANN) recall via vector indexing (e.g., FAISS) with typical response times of 1–5ms for $k=5$ recall at embedding dimensions $d=128$~\cite{johnson2017billion}. In-process or shared-memory deployments can reduce this further to $<1$ms, ensuring that Cortex does not become a bottleneck in xApp inference.

rApps, operating at timescales of seconds or minutes, face no such constraints and can use batch-mode recall or offline analytics. Thus, the Cortex pipeline is feasible for both latency domains without requiring specialized inference hardware.

\subsection{Comparison with Existing Architectures}

Compared to existing xApp/rApp pipelines, Cortex introduces a structured memory access pattern that enables:
\begin{itemize}
    \item Sample-efficient adaptation to unseen scenarios via similarity matching.
    \item Explainable policy recommendations based on historical precedent.
    \item Continuous policy refinement without end-to-end retraining.
\end{itemize}

By contrast, current stateless agents rely entirely on online feature extraction and periodic retraining, which delays convergence and reduces generalization~\cite{arulkumaran2019deep, yang2022ran}.

\subsection{System Compatibility and Deployment Risk}

Cortex is strictly additive and modular: it requires no modification to the E2, F1, or O-RAN fronthaul interfaces. It also preserves all existing rApp/xApp control paths. This design ensures backward compatibility and minimal deployment risk. Should Cortex fail or degrade, policy execution falls back to native stateless logic, preserving operational safety.

\subsection{Expected Benefits and Impact}

We anticipate performance improvements in:
\begin{itemize}
    \item \textbf{Mobility management:} Recall-augmented policies can track multi-cell handover chains and recurrent mobility failures, improving UE retention and handover success.
    \item \textbf{Admission control:} xApps can recall congested historical contexts to adjust thresholds preemptively, improving SLA enforcement.
    \item \textbf{Beamforming:} Agents can exploit structural similarities in propagation environments, enabling faster adaptation to spatial anomalies.
\end{itemize}

In each case, the ability to retrieve actionable priors without retraining presents a compelling trade-off between performance, explainability, and infrastructure cost.

\subsection{Validation Roadmap}

To validate Cortex, we propose:
\begin{itemize}
    \item Integration with an open O-RAN RIC emulator (e.g., ColO-RIC~\cite{coloric2023} or Near-RT RIC from O-RAN SC).
    \item Use of public xApp/rApp benchmarks with synthetic load traces to simulate memory usage and policy improvements.
    \item Side-by-side comparisons with baseline agents using identical network scenarios, evaluating key KPIs such as throughput, call drop rate, and SLA violations.
\end{itemize}

These steps would allow the community to benchmark memory-augmented RIC agents without disrupting existing production workflows, and to validate whether retrieval-augmented learning can be generalized across diverse network topologies.

\section{Related Work}

The application of AI to Radio Access Networks has gained substantial momentum in recent years, with a particular focus on control optimization, anomaly detection, and resource allocation. However, most existing efforts remain stateless, relying on supervised learning or reinforcement learning agents that lack long-term memory mechanisms or contextual recall. This section reviews prior approaches across xApp/rApp design, RIC architecture extensions, and memory-based AI systems.

\paragraph{AI in O-RAN xApps and rApps.}  
Several projects have proposed AI-driven xApps for functions such as traffic steering~\cite{yang2022ran}, power control~\cite{zhang2021ai}, and beamforming optimization~\cite{shlezinger2020viterbinet}. These models typically operate using snapshot RAN telemetry or engineered features. rApps have been explored for mobility prediction and slice resource planning, often with batch-mode retraining via digital twins or network emulators~\cite{filali2022oran}. However, these agents process state data statelessly, without explicit memory augmentation or episodic knowledge reuse.

\paragraph{Retrieval-Augmented Learning.}  
In natural language processing and vision domains, retrieval-augmented models have shown dramatic improvements in generalization, data efficiency, and interpretability~\cite{borgeaud2022retro, izacard2021leveraging}. These models use external vector stores to condition inference on relevant historical samples. To date, no known implementation has applied such retrieval mechanisms to AI-driven RAN decision agents, despite the similarity in state-action structure between RAN control loops and reinforcement learning environments.

\paragraph{O-RAN Platform Extensions.}  
Efforts to enhance RIC functionality have included runtime model orchestration~\cite{oran-security}, inference pipelining~\cite{foukas2023ran}, and closed-loop adaptation using digital twins~\cite{mirza2023dtwran}. However, none of these works have addressed the integration of episodic memory structures or semantic recall layers within the inference path of xApps or rApps. RAN Cortex fills this architectural gap by offering a modular, inference-agnostic recall interface that is compliant with O-RAN modularity.

\paragraph{Memory Systems in Reinforcement Learning.}  
In adjacent AI domains, neural episodic memory systems~\cite{pritzel2017neural, graves2016hybrid} and key-value memory networks~\cite{sukhbaatar2015end} have enabled agents to perform context-aware reasoning in partially observable or data-sparse environments. While these systems require tight coupling with agent architecture, RAN Cortex abstracts memory into an external, interoperable service, enabling practical deployment within production telco environments.

To the best of our knowledge, this is the first proposal to bring retrieval-based memory augmentation into real-time AI-RAN control pipelines, enabling semantically grounded decision support for both xApps and rApps in a standard-compliant manner.

\section{Use Cases}

The RAN Cortex architecture introduces a modular intelligence layer that is applicable across a broad range of near-RT and non-RT RAN functions. This section highlights representative use cases that demonstrate the architectural flexibility and operational benefit of memory-augmented decision-making.

\subsection{Mobility Management under Dynamic Topologies}

In ultra-dense urban scenarios or high-speed vehicular contexts, handover chains often experience instability due to variable signal quality, backhaul delays, or policy misalignment. By recalling past sequences of mobility failures and their outcomes, rApps can dynamically refine handover thresholds and blacklist ineffective target cells. This is particularly useful in networks with shared infrastructure or rapidly reconfigured coverage footprints.

\subsection{Admission Control with SLA Guarantees}

Admission control xApps typically operate under fixed rules or periodically retrained thresholds. Cortex enables these modules to condition access decisions on previously seen traffic mixtures and cell loading scenarios, especially under SLA constraints for enterprise slices or URLLC traffic. Recall of prior outcomes allows for predictive admission rejection that prevents SLA degradation preemptively.

\subsection{Beamforming and Interference Coordination}

xApps handling beam selection and power control can benefit from retrieval of previously observed spatial contexts, enabling more informed beam vector decisions in anomalous propagation environments. This is especially critical in mmWave deployments or when operating under limited channel feedback due to energy or latency constraints.

\subsection{Anomaly Detection and Self-Healing}

Cortex allows retrospective pattern matching for rare but high-impact anomalies, such as random access channel congestion or massive synchronized call drops. By maintaining episodic records of prior faults, rApps can perform rapid root-cause triangulation, even when anomalies do not trigger traditional alarm thresholds.

\subsection{Digital Twin Synchronization and Policy Testing}

In testbed or offline settings, Cortex can simulate memory-guided policy performance under synthetic or replayed network traces. This enables safe experimentation with long-term policy learning and fast adaptation strategies before live deployment.

These use cases illustrate that RAN Cortex is not restricted to a specific control domain. Instead, it generalizes across spatial, temporal, and topological dimensions of network behavior, offering consistent memory-based priors across applications.

\section{Next Steps, Considerations, and Conclusions}

\subsection{Next Steps}

To validate RAN Cortex, our immediate goal is to prototype the architecture using public RIC development environments such as the Near-RT RIC from the O-RAN Software Community or ColO-RIC~\cite{coloric2023}. This prototype will include a lightweight encoder (e.g., temporal CNN or transformer), a pluggable vector store (e.g., FAISS or ScaNN), and a REST/gRPC memory API compatible with existing xApp/rApp control logic.

Key development tracks include:
\begin{itemize}
    \item \textbf{Protocol alignment:} Formalizing schema definitions for recall queries compatible with A1 or RIC internal messaging.
    \item \textbf{Indexing policy:} Designing intelligent storage policies to prioritize high-salience or failure-prone network states.
    \item \textbf{Latency benchmarking:} Measuring retrieval latency across deployment variants (in-process vs. microservice) under different hardware configurations.
    \item \textbf{Safety guarantees:} Ensuring determinism and graceful degradation paths in scenarios where the Cortex layer is degraded or offline.
\end{itemize}

These steps are designed to demonstrate feasibility in near-production environments while providing benchmarking baselines for integration by vendors and operators.

\subsection{Design Considerations}

The architecture of RAN Cortex is deliberately inference-agnostic and modular. It makes no assumptions about policy structure—whether rule-based, supervised, or reinforcement learning—and can serve as an auxiliary advisory module or as a primary input into a composite policy engine. This design ensures compatibility with legacy logic and allows gradual adoption across heterogeneous deployments.

Cortex also supports deployment in privacy-sensitive environments. Embeddings and outcomes can be encrypted at rest, tenant-separated for multi-slice architectures, and logged through operator observability pipelines. The recall interface can be hardened with authentication, access control lists, and timeouts to prevent leakage or inference disruption.

\subsection{Conclusions}

We have introduced \textit{RAN Cortex}, a modular, memory-augmented retrieval engine for decision enhancement in AI-native RANs. RAN Cortex exposes real-time access to semantically similar prior states, enabling xApps and rApps to leverage historical context without retraining or additional model complexity. The architecture is low-friction, compatible with existing O-RAN interfaces, and latency-aligned with near-RT and non-RT control loops.

This proposal represents the first application of retrieval-augmented inference in the context of RAN control agents. By decoupling memory from policy and standardizing recall as a service, RAN Cortex provides a foundation for scalable, explainable, and context-aware optimization across 5G and emerging 6G networks.

The future of AI in telecom will not rely solely on smarter models, but on smarter memory: structured, efficient, and interoperable across the stack. RAN Cortex is a step toward that future.

\end{document}